\documentclass{jpsj3}
\usepackage{txfonts}
\usepackage{graphicx}
\usepackage{color}
\usepackage{mathrsfs}
\usepackage{bm}
\usepackage{multirow}
\usepackage{booktabs}
\usepackage[left]{lineno}
\usepackage[ruled]{algorithm}
\usepackage{algpseudocode}
\usepackage{algpascal}
\usepackage{algc}
\usepackage{url}

\newcommand\ASTART{\bigskip\noindent\begin{minipage}[b]{0.5\linewidth}}

\newcommand\AENDSKIP{\end{minipage}\bigskip}
\newcommand\AEND{\end{minipage}}

\newcommand{\V}[1]{\bm{#1} } 

\newcommand{\mR}{\mathbb{R}}

\newcommand{\lb}{\left(}
\newcommand{\rb}{\right)}
\newcommand{\lbb}{\left\{}
\newcommand{\rbb}{\right\}}

\newcommand{\Req}[1]{eq.\ (\ref{eq:#1})}

\newcommand{\NReq}[1]{(\ref{eq:#1})}

\newcommand{\Rfig}[1]{Fig.\ \ref{fig:#1}}

\newcommand{\Lfig}[1]{\label{fig:#1}}
\newcommand{\Leq}[1]{\label{eq:#1}}
\newcommand{\Rsec}[1]{sec.\ \ref{sec:#1}}
\newcommand{\Lsec}[1]{\label{sec:#1}}
\newcommand{\be}{\begin{eqnarray}}
\newcommand{\ee}{\end{eqnarray}}
\newcommand{\ba}{\begin{array}}
\newcommand{\ea}{\end{array}}

\newcommand{\subbe}{\begin{subequations}}
\newcommand{\subee}{\end{subequations}}
\newcommand{\bs}{\backslash}
\newcommand{\mc}[1]{\mathcal{#1}}
\DeclareMathOperator*{\argmin}{arg\,min}

\newcommand{\nnzero}{K}
\newcommand{\MSEx}{\epsilon_{x}}
\newcommand{\MSEy}{\epsilon_{y}}

\newcommand{\MSEcv}{\epsilon_{\rm CV}}

\newcommand{\lA}{\leftarrow}

\newcommand{\Lcode}[1]{\label{code:#1}}
\newcommand{\Rcode}[1]{Alg.\ \ref{code:#1}}
\newcommand{\nnzeronum}{K}

\newcommand{\TS}{\bm{x}^0}

\title{Reconstructing Sparse Signals via Greedy Monte-Carlo Search}

\author{
Kao Hayashi$^1$, 
Tomoyuki Obuchi$^2$\thanks{obuchi@i.kyoto-u.ac.jp}, 
Yoshiyuki Kabashima$^{3,4}$
}

\inst{
$^1$Graduate School of Sciences and Humanities, Nara Women's University, Nara 630-8506, Japan \\
$^2$Department of Systems Science, Graduate School of Informatics, Kyoto University, Yoshida Hon-machi, Sakyo-ku, Kyoto-shi, Kyoto 606-8501, Japan \\
$^3$Department of Mathematical and Computing Science, Tokyo Institute of Technology, 2-12-1, Ookayama, Meguro-ku, Tokyo 152-8550, Japan\\
$^4$Institute for Physics of Intelligence \& Department of Physics, Graduate School of Science, The University of Tokyo, 7-3-1 Hongo, Bunkyo-ku, Tokyo 113-0033, Japan
}

\abst{
We propose a Monte-Carlo-based method for reconstructing sparse signals in the formulation of sparse linear regression in a high-dimensional setting. The basic idea of this algorithm is to explicitly select variables or covariates to represent a given data vector or responses and accept randomly generated updates of that selection if and only if the energy or cost function decreases. This algorithm is called the greedy Monte-Carlo (GMC) search algorithm. Its performance is examined via numerical experiments, which suggests that in the noiseless case, GMC can achieve perfect reconstruction in undersampling situations of a reasonable level: it can outperform the $\ell_1$ relaxation but does not reach the algorithmic limit of MC-based methods theoretically clarified by an earlier analysis. The necessary computational time is also examined and compared with that of an algorithm using simulated annealing. Additionally, experiments on the noisy case are conducted on synthetic datasets and on a real-world dataset, supporting the practicality of GMC. 
}

\begin{document}
\maketitle

\section{Introduction} \Lsec{Introduction}
Variable selection is an important problem in statistics, machine learning and signal processing. One of the recent attractive topics regarding this study is compressed sensing~\cite{candes2005decoding,candes2006robust,candes2006near,donoho2006compressed}, in which a sparse signal is reconstructed from (linear) measurements with smaller measurement number than the overall signal dimension. If we denote the measurement result as $\V{y}\in \mR^{M}$ and the measurement matrix as $A\in \mR^{M\times N}$, a naive formulation of compressed sensing can be written as
\be
\hat{\V{x}}(\nnzeronum)=\argmin_{\V{x}}\lbb \frac{1}{2}||\V{y}-A\V{x}||_2^2\rbb~{\rm subj.~to}
~ ||\V{x}||_0\leq \nnzeronum,
\Leq{naiveCS}
\ee
where $||\cdot||_0$ denotes the $\ell_0$ norm giving the number of non-zero components and the $\nnzeronum$-sparse vector $\hat{\V{x}}(\nnzeronum)(\in \mR^N)$ becomes our estimator of the true signal $\TS$ embedded in the data generation process. This formulation requires solving discrete optimization problems; therefore, it involves serious computational difficulty when $N$ and $K$ are large~\cite{natarajan1995sparse}. Due to this computational difficulty, certain approximations are needed to treat high-dimensional datasets.

Some representative approximations, such as orthogonal matching pursuit (OMP)~\cite{pati1993orthogonal,davis1994adaptive} and the iteratively reweighted least squares method (IRLS)~\cite{chartrand2008iterative}, are available. OMP is an approximate algorithm to directly solve~\Req{naiveCS} by incrementing the set of used columns of $A$ to approximate $\V{y}$ in a greedy manner, whereas IRLS first relaxes the $\ell_0$ constraint to an $\ell_p$ constraint, where $p>0$, and then recursively solves the least squares problem with coefficients reweighted by the previous least squares solution; IRLS effectively solves \Req{naiveCS} with a small enough $p$. Another more common method to find an approximate solution of \Req{naiveCS} is to relax $||\V{x}||_0$ to $||\V{x}||_1=\sum_{i=1}^N |x_i|$~\cite{tibshirani1996regression,meinshausen2004consistent,banerjee2006convex,friedman2008sparse}. This relaxation converts the problem into a convex optimization problem; hence, a unique solution can be efficiently obtained using versatile solvers. Even with this relaxation, it has been shown that a perfect reconstruction of the signal in the noiseless limit is possible in undersampling situations of a reasonable level~\cite{donoho2009observed,kabashima2009typical}. These findings have motivated using the $\ell_1$ relaxation (LASSO) for a wide range problems and inspired a search for better inference methods enabling a perfect reconstruction with fewer measurements than the $\ell_1$ case with a reasonable computational cost. A successful example of this search is based on the Bayesian framework~\cite{krzakala2012probabilistic}: it showed that a perfect reconstruction is possible with fewer measurements than the $\ell_1$ case with a sufficiently low computational cost when employing the so-called approximate message passing (AMP) algorithm. However, this impressive result is not perfect because prior knowledge about the generation of $\TS$ and $\V{y}$ is required in the Bayesian framework but is not always available in realistic settings, and also its performance is guaranteed only for a certain class of measurement matrices. Solvers have their own advantages and disadvantages, and the appropriate choice depends on the available resources and purpose of the signal processing. Therefore, it is useful to increase the options of such solvers. 

Under these circumstances, some of the authors proposed a solver based on Monte-Carlo (MC) sampling methods. The solver employs simulated annealing (SA)~\cite{kirkpatrick1983optimization}, which is a versatile metaheuristic for optimization, and its performance was examined numerically and analytically~\cite{obuchi2018statistical,nakanishi2016sparse,obuchi2016sparse,obuchi2016sampling}. The analytical result shows that a fairly wide parameter region exists in which the phase space structure is rather simple and a perfect reconstruction is possible; the existence limit of this region defines the algorithmic limit to achieve a perfect reconstruction and this limit is shown to be comparable with that of the Bayesian approach~\cite{obuchi2018statistical}. Therefore, the SA-based algorithm, which is applicable for any measurement matrices, can exhibit comparable performance with a Bayesian method even without prior knowledge of the generative processes of $\TS$ and $\V{y}$, although its computational cost is higher than that of AMP.

The above result, including the larger computational cost of the SA-based algorithm and the simple structure of the phase space clarified by the analytical computation, inspires a further simplification of the MC-based algorithm, which is the goal of this study. In particular, we propose a greedy algorithm to solve \Req{naiveCS} based on MC sampling, and call it the {\it greedy Monte-Carlo search} (GMC) algorithm. GMC can also be regarded as a version of the SA-based algorithm quenched to zero temperature. Below we examine the performance of GMC via numerical experiments. The result suggests that the GMC performance does not reach that of the SA-based method but is better than LASSO.

\section{Formulation and algorithm} \Lsec{Algorithm}
Let us suppose a data vector $\V{y}\in \mR^{M} $ is generated from the measurement matrix $A\in \mR^{M\times N}$ and a $K_0$--sparse signal $\TS\in \mR^{N},||\TS||_0=K_0$ via
\be
\V{y}=A\TS . \label{eq:y_Ax}
\ee
We aim to infer the signal $\TS$ from a given $A$ and $\V{y}$ for the underdetermined situation $M<N$. We denote the estimated signal as $\V{\hat{x}}$. To solve this underdetermined problem, we assume that the number of non-zero components $K_0=||\TS||_0$ is smaller than or equal to $M$, where $K_0\leq M$. 

The starting point of our inference framework is the least squares method. To apply this method to the undetermined situation, we explicitly specify the variables or columns of $A$ used to represent $\V{y}$ by a binary vector $\V{c}\in \mR^{N}$: if the $i$th variable or column of $A$ is used, then $c_i$ is one; otherwise, it is zero. We call $\V{c}$ the sparse weight. For any $N$--dimensional vector $\V{z}\in \mR^{N}$, the components whose corresponding sparse weights are unity (zero) are called active (inactive) components; we denote the vector of active components as $\V{z}_{\V{c}}$, and that of inactive ones as $\V{z}_{\bar{\V{c}}}$ ($\bar{\V{c}}=1-\V{c}$). Similarly, the submatrix of $A$ consisting only of the active components is denoted as $A_{\V{c}}$. If $K=\sum_ic_i$ is smaller than or equal to $M$, the following least squares estimator is unambiguously defined:
\subbe
\be
&&
\hat{\V{x}}_{\V{c}}(\V{c})
=
\argmin_{\V{x}_{\V{c}}\in \mR^{K}} \lbb \frac{1}{2}||\V{y}-A_{\V{c}}\V{x}_{\V{c}}||_2^2 \rbb
=({A}^{\top}_{\V{c}}A_{\V{c}})^{-1}{A}^{\top}_{\V{c}}{\bm{y}},
\\ &&
\hat{\V{x}}_{\bar{\V{c}}}(\V{c})=\V{0}.
\ee
\Leq{xhat}
\subee
The full dimensional form of this solution is denoted as $\hat{\V{x}}(\V{c})$. Using this least squares estimator, the original problem \NReq{naiveCS} can be represented as
\be
&&
\hat{\V{x}}(K)
=
\hat{\V{x}}(\hat{\V{c}}),
\\ &&
\hat{\V{c}}
=\argmin_{\V{c}:\sum_i c_i = K}
\left\{ \frac{1}{2}||\V{y}-A_{\V{c}}\V{\hat{x}}_{\V{c}}(\V{c})||_2^2 \right\}.
\Leq{combinatorial}
\ee
Therefore, we can solve the problem \NReq{naiveCS} by solving the combinatorial optimization problem \NReq{combinatorial} with respect to $\V{c}$. To solve this problem, the SA-based algorithm was proposed in~\cite{obuchi2016sparse}. Although the SA-based algorithm has been shown to be effective~\cite{obuchi2018statistical,obuchi2016sparse,obuchi2016sampling}, herein we examine a further simplified approach. In particular, we propose a simpler algorithm that solves \Req{combinatorial} in a greedy manner using MC sampling. We call this algorithm GMC, and the details are given in the following subsection. For convenience, we introduce a mean-square error (MSE) when representing $\V{y}$ such that 
\be
\MSEy(\V{c}|\V{y},A)=\frac{1}{2M}||\V{y}-A_{\V{c}}\V{\hat{x}}_{\V{c}}(\V{c})||_2^2,
\ee
and call this the output MSE. This plays the role of `energy' in the MC update as shown below. 

\subsection{GMC} \Lsec{GMC}
The outline of GMC is as follows. Starting from a given initial configuration or state of $\V{c}$, we update the state in an MC manner. We randomly generate a new state $\V{c}'$ and update it as $\V{c}\to \V{c}'$ if and only if the output MSE or energy decreases. If the energy cannot be decreased by any single `local flip', then the algorithm stops and returns the current state as the output. 

Comprehensively, during the update, we require the number of non-zero components $K=\sum_i c_i$ to remain constant. To efficiently generate a new state satisfying this requirement, we employ the `pair flipping' of two sparse weights: one equals $0$ and the other equals $1$. In particular, we randomly choose an index $i$ from ${\rm ONES} \equiv\{k|c_k=1\}$ and another index $j$ from ${\rm ZEROS}\equiv\{k|c_k=0\}$ and we set $(c_i',c_j')=(0,1)$ while keeping the other components unchanged. The pseudocode for pair flipping is given in \Rcode{MC}. 
\alglanguage{pseudocode}
\begin{algorithm}[htbp]
\caption{MC update with pair flipping}\Lcode{MC}
\begin{algorithmic}[1]
\Procedure{MCpf}{$\V{c},\V{y},A$}\Comment{MC routine  with pair flipping}
	\State ${\rm ONES} \lA \{k|c_k=1\},~{\rm ZEROS} \lA \{k|c_k=0\}$
	\State randomly choose $i$ from ONES and $j$ from ZEROS
	\State $\V{c}' \lA \V{c}$
	\State $(c'_i, c'_j) \lA (0,1)$
	\State $(\MSEy,\MSEy')\lA (\MSEy(\V{c}|\V{y},A),\MSEy(\V{c}'|\V{y},A))$ \Comment{Calculate energy}	
	\If {$ \MSEy' < \MSEy $}
		\State $\V{c} \lA \V{c}'$
		\State $\MSEy \lA \MSEy'$
	\EndIf       
	\State \Return $\V{c},\MSEy$
\EndProcedure
\end{algorithmic}
\end{algorithm}
We define one MC step (MCS) as $N$ pair-flipping trials. 

Pair flipping gives a concrete meaning to the `local flip' mentioned above. Accordingly, we can introduce the stopping condition for GMC as follows. If the configuration of $\V{c}$ is invariant during $t_{\rm wait}$ MCSs, we examine all states accessible via a single pair flip from the current state $\V{c}$ and compute the associated energy values; if there is no lower energy state than the current one, then the algorithm stops; otherwise, the state is updated to the lowest energy state of the locally accessible ones and the usual update is continued. The pseudocode for the entire GMC procedure is given in \Rcode{GMC}.
\begin{algorithm}[htbp]
\caption{Greedy Monte-Carlo Search (GMC)}\Lcode{GMC}
\begin{algorithmic}[1]
\Procedure{GMC}{$\V{c}^{\rm ini},\V{y},A,t_{\rm wait}$} 
\State $\V{c} \lA \V{c}^{\rm ini}$, $(M,N) \lA {\rm size}(A)$, $K \lA \sum_{i}c_{i}$, $t\lA 0$ \Comment{Initialization}
\While{}
    \State $\V{c}_{\rm pre} \lA \V{c}$
    \For {$i=1:N$}                \Comment{One MCS}
        \State $\lb \V{c},\MSEy\rb \lA {\rm MC_{PF}}(\V{c},\V{y},A)$  
    \EndFor
    \If{$\V{c}=\V{c}_{\rm pre}$}                       \Comment{State update check}
        \State $t\lA t+1$
    \Else
        \State $t\lA 0$
    \EndIf
    \If{$t\geq t_{\rm wait}$}     \Comment{Exhaustive search of locally accessible states}
        \State Set $\mc{E}$ as a $(K(N-K))$-dim vector, and $C$ as a $N\times (K(N-K))$-dim matrix
        \State $a \lA 1$          \Comment{Index for locally accessible states}
            \For{$i=1:K$}
                \State $k \lA {\rm ONES}(i)$      \Comment{$i$th component of ${\rm ONES}$}
                \For{$j=1:(N-K)$}
                    \State $\V{c}' \lA \V{c}$
                    \State $l \lA {\rm ZEROS}(j)$ \Comment{$j$th component of ${\rm ZEROS}$}
                    \State $(c_k', c_l') \lA (0,1)$ \Comment{Pair flipping}
                    \State $C(:,a) \lA \V{c'}$, $\mc{E}(a) \lA \MSEy(\V{c'} | \V{y}, A)$, 
                    \State $a \lA a+1$
                \EndFor
            \EndFor
        \If{No component of $\mc{E}$ is lower than $\MSEy$}
            \State{Break}            \Comment{Break the while loop}
        \Else
            \State $a^*=\argmin_{a}{\mc{E}(a)}$ \Comment{Choose one randomly if multiple minimums exist}
            \State $\MSEy\lA \mc{E}(a^*)$, $\V{c} \lA C(:,a^*)$
            \State $t\lA 0$
        \EndIf
    \EndIf
\EndWhile
\State \Return $\V{c}$ 
\EndProcedure
\end{algorithmic}
\end{algorithm}
Typically, we set $t_{\rm wait}=10$ in the following experiments. 

If we denote the necessary MCSs until convergence as $N_{\rm conv}$, the scaling of the total computational cost of GMC is $O( K(N-K)N_{\rm energy}+N_{\rm conv}NN_{\rm energy})$, where the first term is for the search of locally accessible states and the last term is for the MC update; $N_{\rm energy}$ is the computational cost of the energy, which is estimated as $O(K^3+MK^2)$ when a naive method of matrix inversion is used but can be reduced to $N_{\rm energy}=O(K^2+MK)$ by using the technique mentioned in~\cite{obuchi2016sparse}. Certainly, for the scaling, the computational cost of GMC is in the same order as that of the SA-based algorithm; however, GMC is actually faster because it does not need the annealing procedure required in the SA-based algorithm, reducing the computational cost by a factor of $O(1)$. In the next section, we examine the actual behaviour of GMC using numerical experiments. The scaling of $N_{\rm conv}$ will also be examined.

\section{Numerical experiments}\Lsec{Numerical}
Herein, the performance of the GMC algorithm is numerically examined. Both simulated and real-world datasets are used. 

\subsection{Simulated dataset}\Lsec{On simulated}
In this subsection, we examine the performance of GMC on simulated datasets, particularly focusing on whether a perfect reconstruction of $\TS$ is achieved. To more directly quantify the reconstruction accuracy of $\TS$, we introduce another MSE in addition to $\MSEy$:
\be
\MSEx(\V{c})=\frac{1}{2N}||\TS-\V{\hat{x}}(\V{c})||_2^2,
\ee
which is referred to as the input MSE hereafter.

Our simulated datasets are generated as follows. Each component of the design matrix $A\in \mR^{M\times N}$ is independent and identically distributed (i.i.d.) from $\mathcal{N}\lb 0,N^{-1} \rb$; the non-zero components of the $K_0$--sparse signal $\TS\in \mR^{N}$ are also i.i.d. from $\mathcal{N}(0,1/\rho_0)$, where $\rho_0=K_0/N$ is the density of the non-zero components, setting the power of the signal to unity. The data vector is then obtained by \Req{y_Ax} given $A$ and $\TS$. This setup is identical to that of the theoretical computation in~\cite{obuchi2018statistical}; thus, we can directly compare the results. We follow the limit of theoretical computation, where the thermodynamic limit $N\to \infty$ is considered while maintaining the ratios $\alpha=M/N,\rho=K/N,\rho_0=K_0/N$ in $O(1)$.

\subsubsection{Noiseless case}\Lsec{Noiseless case}
Herein, we examine the perfect reconstruction ratio obtained by GMC when the sparsity is correctly specified $\rho=\rho_0$. In particular, we prepare $N_{\rm init}$ initial conditions with the correct sparsity $\rho=\rho_0$ for each sample of $(\TS,A)$, run GMC from the initial conditions and compute the ratio achieving a perfect reconstruction of the true signal. We call this ratio the {\it success rate} ($P_{\rm suc}$). The ensemble of $P_{\rm suc}$ for $N_{\rm samp}$ samples quantifies the performance of GMC and is therefore investigated below. For simplicity, we fix $N_{\rm init}=100$ in the following experiments. 

The average value of $P_{\rm suc}$ over $N_{\rm samp}=100$ is listed against the system size in Table \ref{table:p_cinit}.   
\begin{table}[h]
\begin{center}
\begin{tabular}{|c||c|c|c|c|c|} \hline
  $N$             &  100          &  200           &  400          &  800  &  1000 \\ \hline
$P_{\rm suc}$ & 0.56$\pm0.02$ &  0.54$\pm0.02$ & 0.52$\pm0.02$ &  0.67 $\pm$0.07   & 0.75 $\pm$ 0.04        \\ \hline
\end{tabular}
\caption{The system size dependence of the average value of $P_{\rm suc}$. The dependence seems to be absent. }
\label{table:p_cinit}
\end{center}
\end{table}
The examined system sizes are $N=100, 200, 400, 800,1000$ and the other parameters are fixed to $\alpha=0.5, \rho_0=\rho=0.2$. The error bar is the standard error over the $N_{\rm samp}$ samples. The result implies that a system size dependence is very weak or absent and GMC can stably find the true solution.

Further, we investigate the ratio of samples exhibiting $P_{\rm suc}>0$, $P_{\rm samp}$, for a wide range of parameters $(\alpha,\rho_0)$. This allows us to capture the practical algorithmic limit of the perfect reconstruction by GMC. A heat map of $P_{\rm samp}$ for the overall region of $(\alpha,\rho_0)$ at $N=100$ is plotted in the left panel of \Rfig{Psamp}.   
\begin{figure}[htbp]
\begin{center}
\includegraphics[width=0.45\columnwidth]{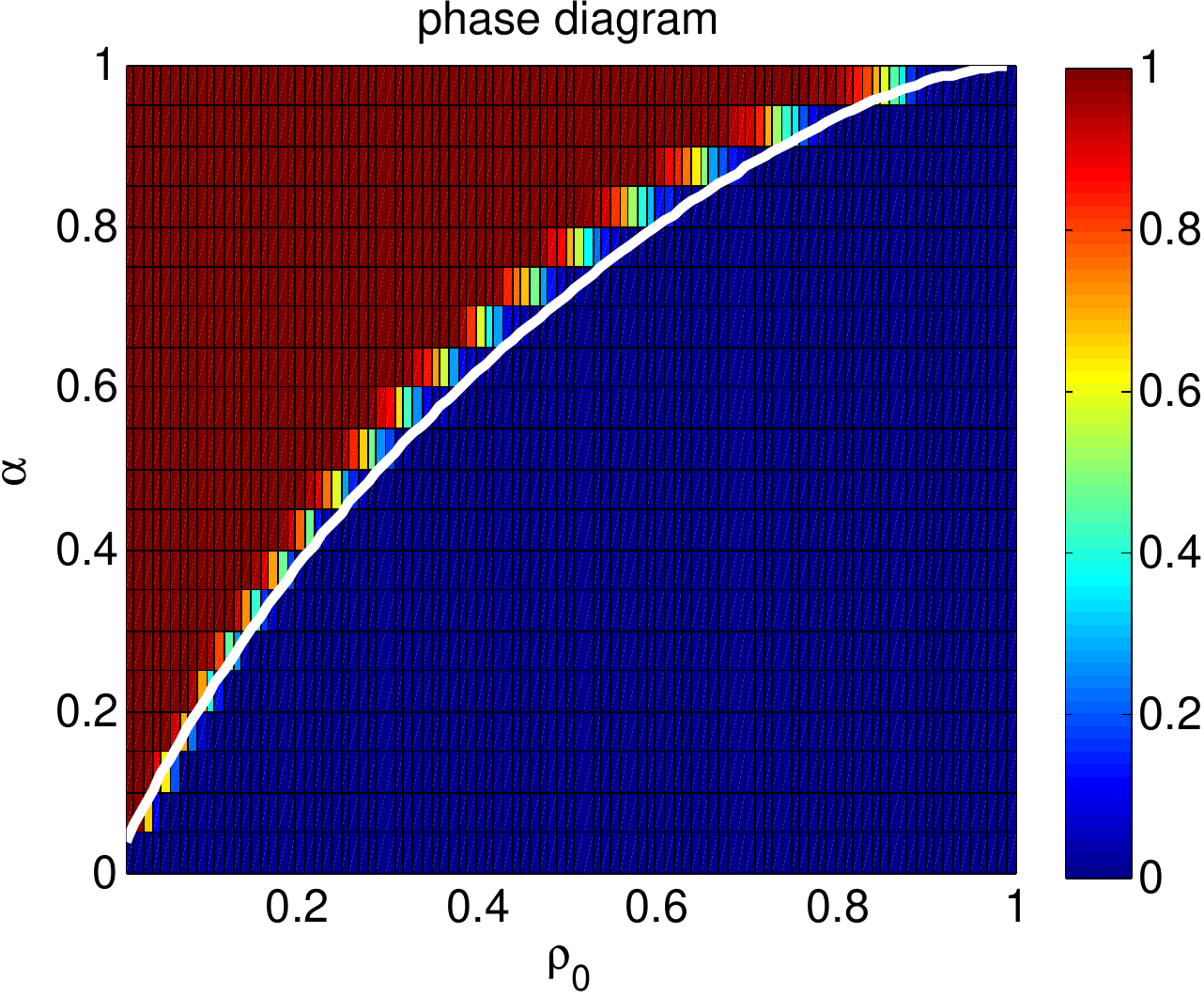}
\includegraphics[width=0.45\columnwidth]{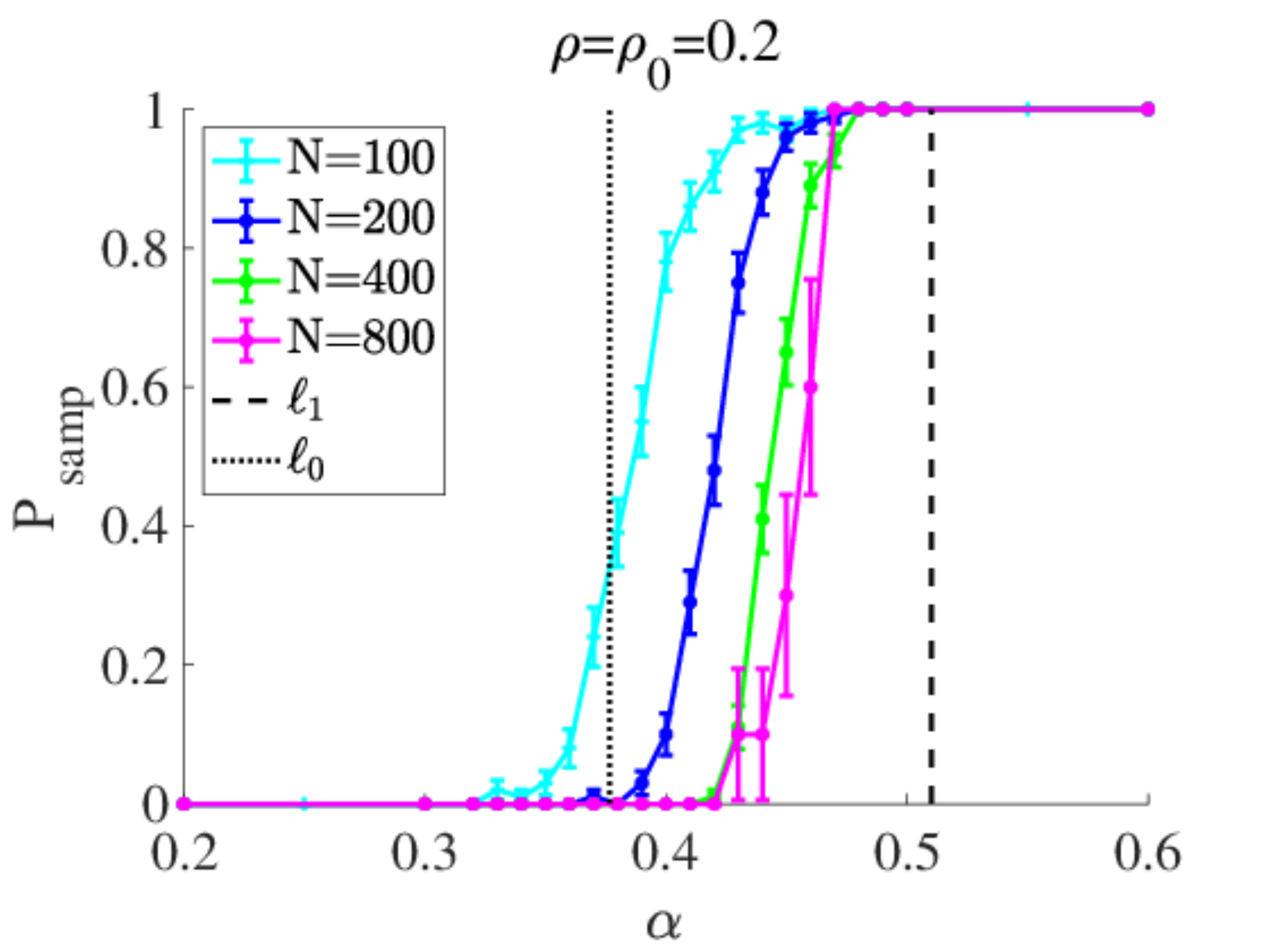}
\caption{(Left) A heat map of $P_{\rm samp}$ at $N=100$ for the overall range of the parameters $(\alpha,\rho_0)$. The while line is the theoretically derived limit obtained in~\cite{obuchi2018statistical}. (Right) Plots of $P_{\rm samp}$ against $\alpha$ at $\rho=0.2$ for different system sizes. The black dotted line ($\ell_0$) denotes the limit derived in~\cite{obuchi2018statistical} while the black dashed line ($\ell_1$) is the theoretical limit of the $\ell_1$ relaxation~\cite{donoho2009observed,kabashima2009typical}.}
\Lfig{Psamp}
\end{center}
\end{figure}
Herein, $P_{\rm samp}$ is computed for $N_{\rm samp}=100$ samples. The theoretically derived algorithmic limit is depicted by a white line, and a sharp change of the colour occurs around the line. This implies that GMC achieves the limit. To examine this point, we further investigate the system size dependence of $P_{\rm samp}$. The result is indicated in the plot of $P_{\rm samp}$ against $\alpha$ at $\rho_0=0.2$ in the right panel of the same figure. As $N$ grows, the curve of $P_{\rm samp}$ becomes steeper, implying the presence of phase transition. Although the precise location of the transition point is unclear, it seems to be larger than the algorithmic limit (black dotted line) derived in~\cite{obuchi2018statistical}. This unfortunately means that the GMC performance is worse than the SA-based method. Meanwhile, the transition point location seems to be still smaller than the reconstruction limit of the $\ell_1$ relaxation (black dashed line), implying that GMC outperforms LASSO and thus can be a reasonable solver. 

Thirdly, we list the necessary MCSs until convergence in \Rfig{N_conv}. 
\begin{figure}[htbp]
\begin{center}
\includegraphics[width=0.45\columnwidth]{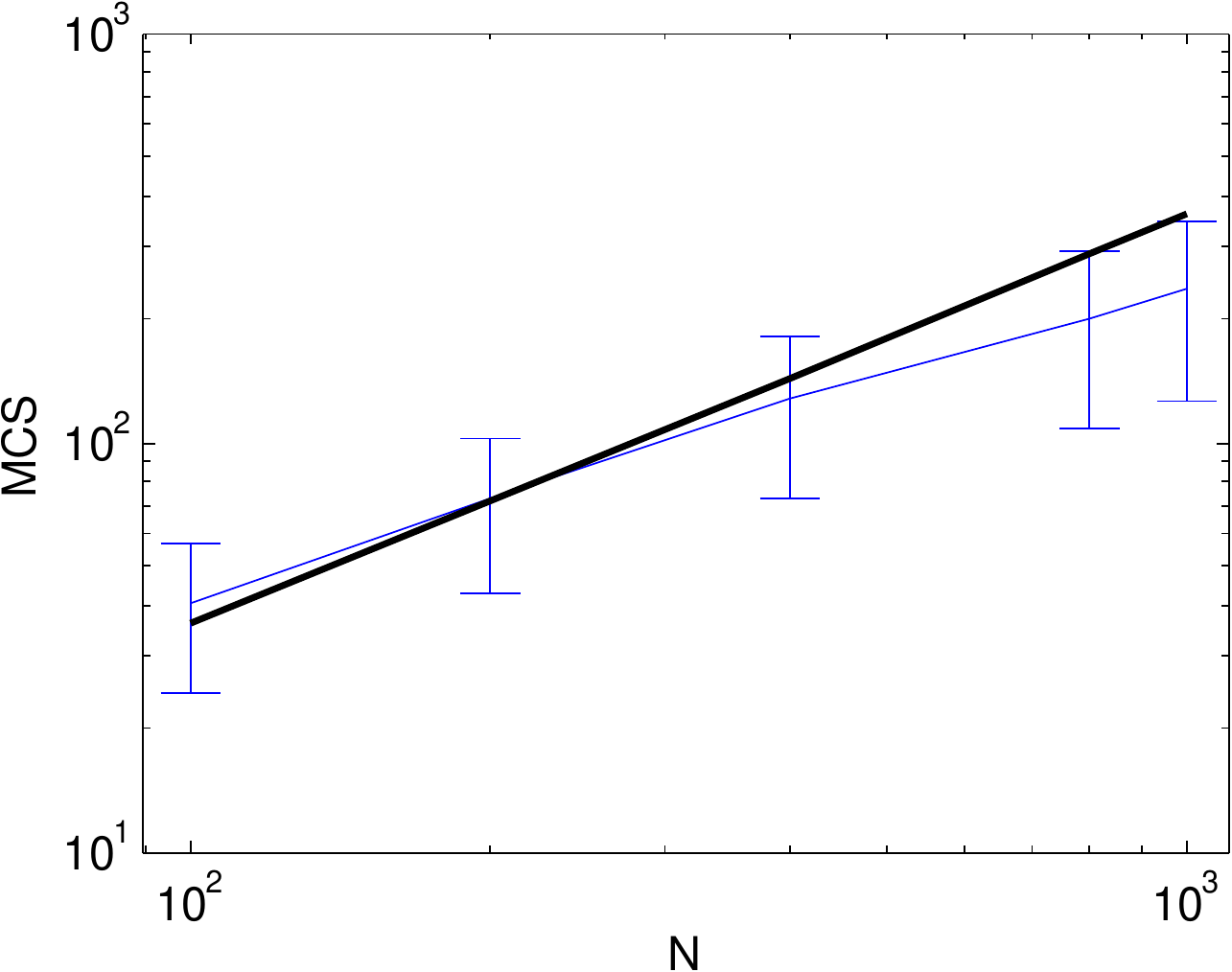}
\caption{The system size dependence of $N_{\rm conv}$. A linear increasing tendency with respect to $N$ is observed.}
\Lfig{N_conv}
\end{center}
\end{figure}
Herein, the average value over $N_{\rm samp}=100$ is shown with the standard error. The result implies a linear increase in $N_{\rm conv}$ with respect to $N$. Therefore, the total computational cost of GMC is scaled as $O( K(N-K)N_{\rm energy}$$+N^2N_{\rm energy} )$. This is certainly not cheap; however, GMC can achieve a perfect reconstruction even for rather large values of $\rho_0$, which is not possible using other greedy algorithms such as OMP. This again suggests that GMC is a reasonable choice for sparse linear regression.

Finally, we compare the actual computational time between GMC and the SA-based method. The result is plotted against the system size in \Rfig{comptime}. In the SA algorithm, we adopt the schedule of \cite{obuchi2018statistical}: $N_{T}=200$ temperature points, the highest and lowest temperatures of which are $T=50$ and $10^{-6}$, are selected adaptively combining the arithmetic and geometric progression way; the waiting time $\tau$ at each temperature is fixed to be $\tau=5$.
\begin{figure}[htbp]
\begin{center}
\includegraphics[width=0.45\columnwidth]{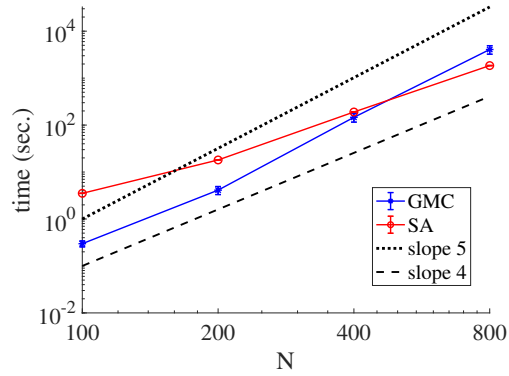}
\caption{Plot of actual computational time of GMC and the SA-based algorithm against the system size. The other parameters are $(\rho,\rho_0,\alpha)=(0.2,0.2,0.5)$ where the perfect reconstruction is possible by both the algorithms, and we checked it is actually achieved in these simulations. Slopes of $5$ and $4$ correspond to the expected scaling of computational time for GMC and SA, respectively. For both the algorithms, $10$ different samples are examined for all the system sizes, and the error bar is obtained as the standard error among the samples. }
\Lfig{comptime}
\end{center}
\end{figure}
In both the algorithms, we commonly use the naive method of matrix inversion in the energy computation, and thus $N_{\rm energy}=O(K^3+MK^2)=O(N^{3})$. In one MCS, $N$ trial flips are conducted and thus the total computational cost of the SA algorithm is expected to be scaled as $O(N^4)$. Meanwhile, based on the discussion in \Rsec{GMC} and the finding in \Rfig{N_conv}, the total computational cost of GMC is expected to be $O(N^5)$. These scalings are observed in \Rfig{comptime}. This seems to imply that GMC is worse than the SA algorithm, but we have two noteworthy remarks against this implication. The first point is that at least GMC is faster at small $N$ as observed in \Rfig{comptime}, which would be due to the SA's extra computational cost coming from simulating many temperature points. The second point is that the SA's scaling $O(N^4)$ is not completely reliable: this scaling is derived assuming the necessary temperature points $N_{T}$ and the waiting time $\tau$ for perfect reconstruction are independent of the system size $N$, but this assumption has no strong reasoning. In the examined range of the system sizes, $(N_{T},\tau)=(200,5)$ seems to be sufficient, but for larger systems, this may not the case. To see the true scaling necessary for the perfect reconstruction by the SA algorithm, further detailed investigations should be necessary but we leave it as a future work.

\subsubsection{Noisy case}\Lsec{Noisy case}
Here we examine the performance of GMC on the noisy case. The data vector is assumed to be generated through 
\be
\V{y}=A\V{x}^{0}+\V{\xi},
\ee
where $\V{\xi}$ is the noise vector each component of which is i.i.d. from $\mathcal{N}(0,\sigma_{\xi}^2)$. The reconstruction performance of $\V{x}^0$ is the subject of interest and we quantify this by $\MSEy$ and $\MSEx$.

To provide a solid advance, we compare the GMC result with the theoretical and simulation results in~\cite{obuchi2018statistical}. In \Rfig{noisy}, we plot $\MSEy$ and $\MSEx$ achieved by GMC against $\rho$ at $(\rho_0,\alpha,\sigma_{\xi}^2)=(0.2,0.5,0.1)$ for different system sizes. Here, for each sample of $\{\V{y},\V{A}\}$, GMC tries $N_{\rm init}=100$ different initial conditions and the one with the smallest $\MSEy$ is adopted as our estimator. Each data point in the plot is obtained from the experiments over $N_{\rm samp}$ independent samples, where $N_{\rm samp}=100$ and $20$ are taken for $N=100$ and $200$, respectively.
\begin{figure}[htbp]
\begin{center}
\includegraphics[width=0.45\columnwidth]{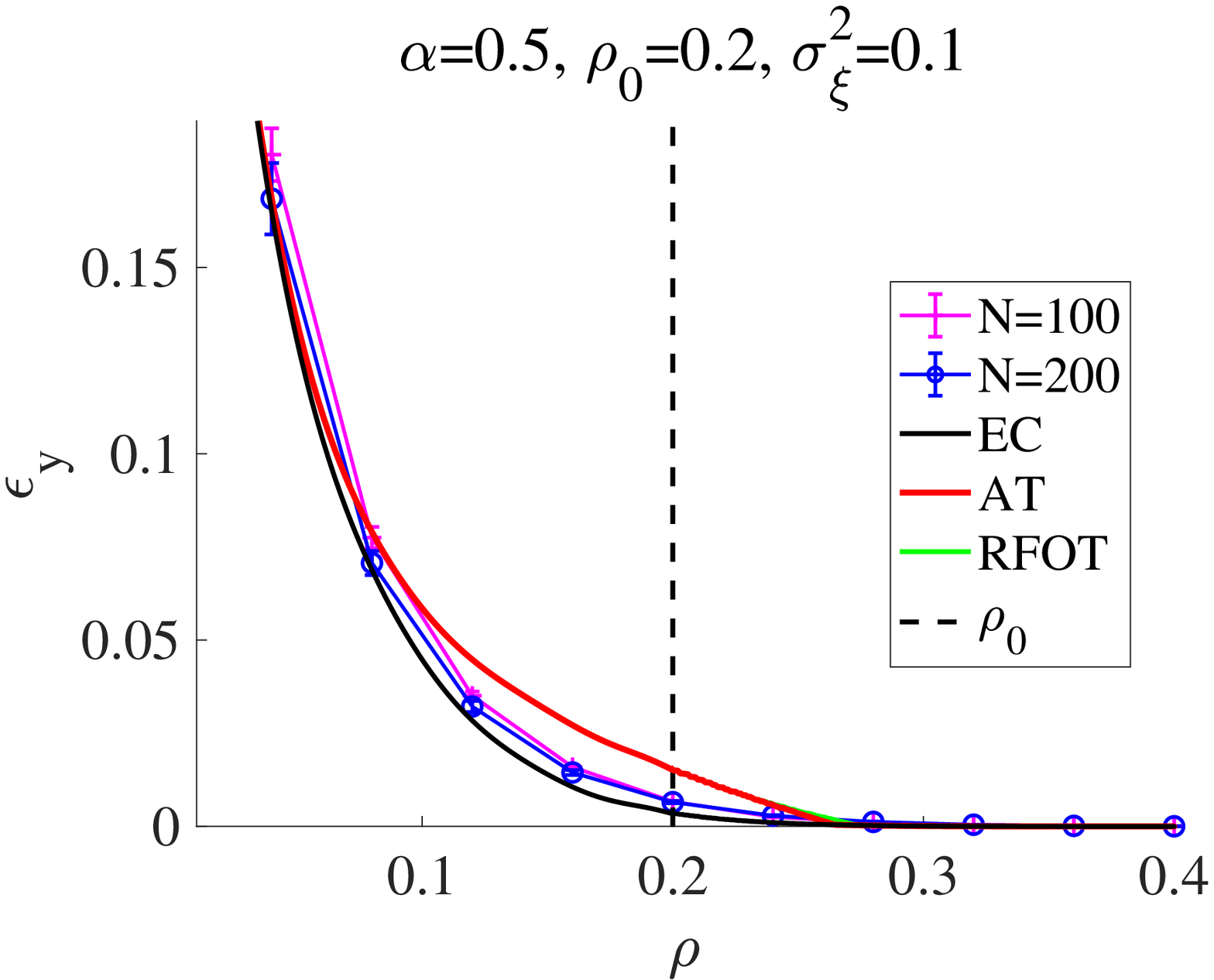}
\includegraphics[width=0.45\columnwidth]{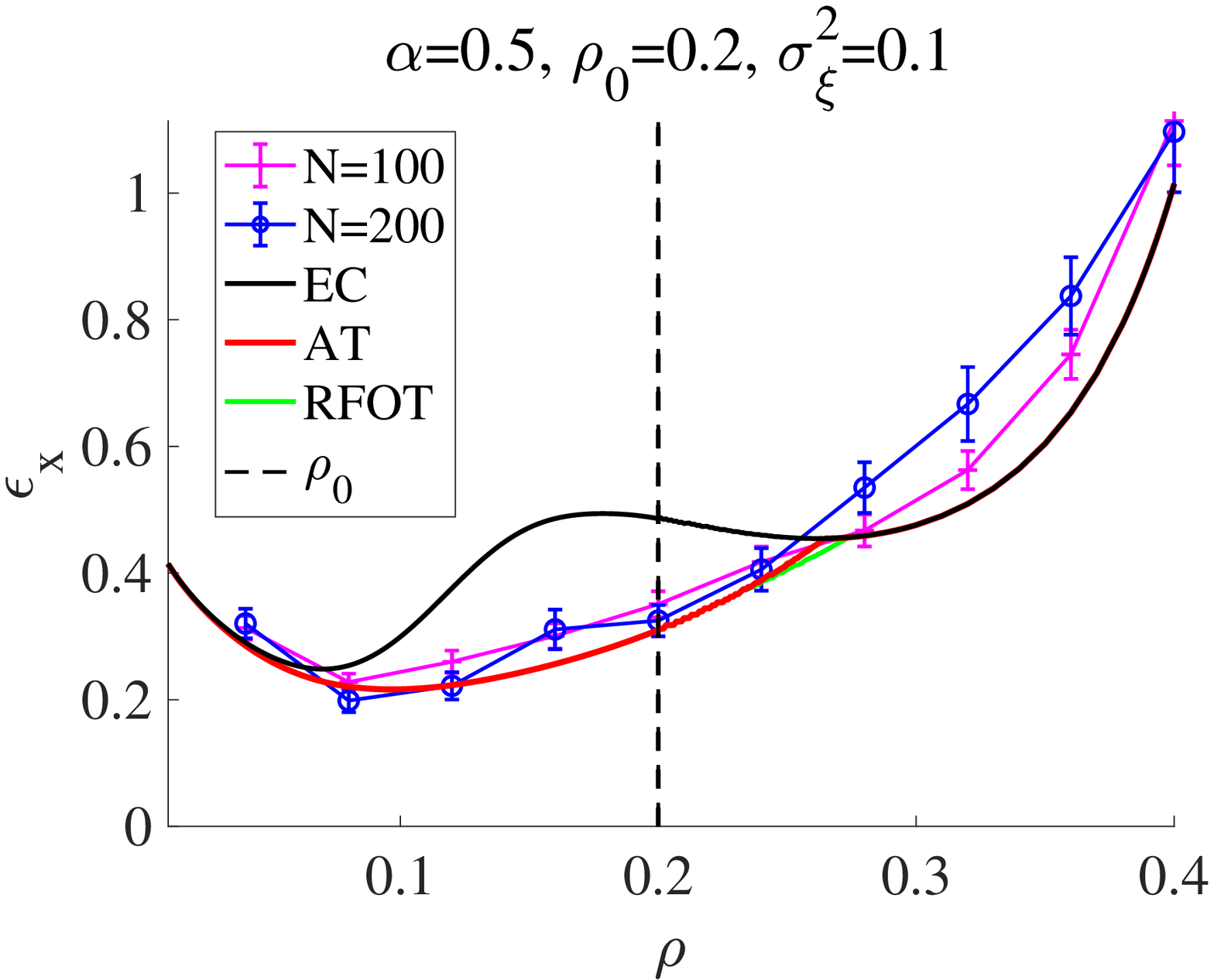}
\caption{Plot of $\MSEy$ (left) and $\MSEx$ (right) against $\rho$ for $N=100,200$ at $\sigma_{\xi}^2=0.1$. The other parameters are $(\rho_0,\alpha)=(0.2,0.5)$. Numerical results (color markers) are compared with the theoretical limits and are found to be matched well with the spin-glass transition lines (AT and RFOT). For the details, see the main text and~\cite{obuchi2018statistical}. Figure 14 in~\cite{obuchi2018statistical} are the counterpart of this figure.  }
\Lfig{noisy}
\end{center}
\end{figure}
We draw several theoretical curves: the black one (EC) for the entropy crisis in the RS level, red (AT) and green (RFOT) ones for the spin-glass transitions of two different natures. For the details, see~\cite{obuchi2018statistical}. Basically, these spin-glass transition lines play as a role of lower bounds of errors achievable by the SA-based algorithm. The GMC result well matches with these lines, and the SA result does so as shown in~\cite{obuchi2018statistical}. This implies that GMC can exhibit comparable performance with the SA algorithm even in the noisy case. This supports the application of GMC to real-world datasets. One such example is shown in the next subsection.

\subsection{Real-world dataset}\Lsec{Real-world dataset}
Herein, we apply GMC to a dataset of Type Ia supernovae. Our dataset is a part of the data from \cite{silverman2012berkeley,Berkeley}, which was screened using a certain criterion~\cite{uemura2015variable}. This dataset was recently treated using a number of sparse estimation techniques, and a set of important variables known to be empirically significant has been reproduced~\cite{obuchi2016sampling,uemura2015variable,obuchi2016cross,kabashima2016approximate}. In these studies, LASSO and $\ell_0$ cases were treated and cross-validation (CV) was employed for the hyperparameter estimation. We reanalyse this dataset using GMC. The parameters of the screened data are $M=78$ and $N=276$, and a standardization, i.e. centering $\V{y}$ and columns of $A$ and normalizing the columns of $A$ to be unit norm, was employed in the pre-processing.

Table \ref{table:supernovae} shows the GMC result for the supernovae dataset. Herein, we changed the value of $K$ from $1$ to $5$, and for each $K$ we examined ten different initial conditions and picked the lowest energy configuration among the tens as our final estimator $\hat{\V{c}}$. The result implies that the output MSE decreases as $\nnzero$ grows, and some variables such as ``2'' and ``1'' seem to be frequently selected.
\begin{table}[h]
\begin{center}
\begin{tabular}{|c||c|c|c|c|c|} \hline
  $K$               &       1     &      2       &       3        &       4      &        5       \\ 
\hline
$\MSEy$         &      0.0312      &    0.0204       &   0.0179        &    0.0158     &   0.0139     \\ 
\hline
variables       &    \{2\}       &   \{1,2\}       &  \{1,2,233\}      &  \{1,2,94,233\}    &    \{2,6,170,221,225\}    \\ 
\hline
\end{tabular}
\caption{The output MSE $\MSEy$ and selected variables' indices for the Type Ia supernovae dataset from~\cite{uemura2015variable} obtained by GMC. The final estimator of $\V{c}$ is the least $\MSEy$ configuration obtained by ten independent runs of GMC with different initial conditions.}
\label{table:supernovae}
\end{center}
\end{table}

To scrutinize this result, we also computed the leave-one-out (LOO) cross-validation (CV) error:
\be
\MSEcv\lb K \biggl( =\sum_{i}\hat{c}_i^{\bs \mu},\forall{\mu} \biggr) \Biggr| \V{y},A \rb  =\frac{1}{2M}\sum_{\mu=1}^{M}\lb y_{\mu}-\sum_{i=1}^N
A_{\mu i}x^{\bs \mu}_{i}(\hat{\V{c}}^{\bs \mu})\rb^2,
\Leq{LOOE}
\ee
where $\V{x}^{\bs \mu}(\V{c})$ is the solution of \Req{xhat} given $\V{c}$ and ``$\mu$th LOO system $\{A^{\bs \mu},\V{y}^{\mu}\}$'': $\V{y}^{\mu}$ is the data vector whose $\mu$th component is removed from the original one as $\V{y}^{\mu}=(y_1,y_2,\cdots,y_{\mu-1},y_{\mu+1},\cdots,y_M)^{\top}$, and $A^{\bs \mu}$ is defined similarly. In our present case, the $\mu$th LOO estimator $\hat{\V{c}}^{\bs \mu}$ is computed by running GMC for the $\mu$th LOO system, and thus can be different for each LOO system. 

The result of a typical single run of GMC is shown in Table~\ref{tab:supernova_LOOE}. Here, the LOO CV error takes its minimum at $K=2$, implying that the best model is obtained at this sparsity level. For further quantification of statistical correlations between variables, we count how many times each variable was selected in this LOO CV procedure, following the way of~\cite{obuchi2016sampling}. Table \ref{tab:supernova_count} summarizes the results for five variables from the top for $\nnzero=1$--$5$. This indicates that no variables other than ``2'' representing {\em light width} were chosen stably, whereas variable ``1'' representing {\em color} was selected with high frequencies for $\nnzero \le 4$. Table \ref{tab:supernova_count} shows that the frequency of ``1'' being selected is significantly reduced for $K=5$. This is presumably due to the strong statistical correlations between ``1'' and the newly added variables. In addition, the results for $\nnzero \geq 4$ varied as we rerun GMC with different initial conditions. These observations mean that we could select at most only {\em light width} and {\em color} as the relevant variables. This conclusion is consistent with those of recent papers~\cite{obuchi2016sampling,uemura2015variable,obuchi2016cross,kabashima2016approximate} using a number of methods including the SA-based one. The similar conclusion is obtained by further detailed analysis using the MC method~\cite{igarashi2018exhaustive,igarashi2018es-dos:}. This provides additional evidence for the practicality of the GMC algorithm.
\begin{table}
\small
\begin{center}
\begin{tabular}{|l ||c|c|c|c |c| }
\hline
$\nnzero$ & 1 & 2 & 3 & 4 & 5 \\
\hline
$\MSEcv$ & 0.0328& 0.0239 &  0.0261 & 0.0302 & 0.0342 \\
\hline
\end{tabular}
\caption{\label{tab:supernova_LOOE}
LOO CV error obtained for $\nnzero=1$--$5$ for the type Ia supernova data set. }
\end{center}
\vspace*{-0.3cm}
\end{table}
\begin{table}[htbp]
\begin{center}
\begin{tabular}{|c|l|lllll|}
\hline
\multicolumn{1}{|l|}{K} &                &                         &                         &                          &                        &   
\\ \hline \hline
\multirow{2}{*}{1}      & variables      & \multicolumn{1}{l|}{2}  & \multicolumn{1}{l|}{*}  & \multicolumn{1}{l|}{*}   & \multicolumn{1}{l|}{*} & * \\ \cline{2-7} 
                        & times selected & \multicolumn{1}{l|}{78} & \multicolumn{1}{l|}{0}  & \multicolumn{1}{l|}{0}   & \multicolumn{1}{l|}{0} & 0 \\ \hline \hline
\multirow{2}{*}{2}      & variables      & \multicolumn{1}{l|}{2}  & \multicolumn{1}{l|}{1}  & \multicolumn{1}{l|}{275} & \multicolumn{1}{l|}{*} & * \\ \cline{2-7} 
                        & times selected & \multicolumn{1}{l|}{78} & \multicolumn{1}{l|}{77} & \multicolumn{1}{l|}{1}   & \multicolumn{1}{l|}{0} & 0 \\ \hline \hline
\multirow{2}{*}{3}      & variables      & \multicolumn{1}{l|}{2}   & \multicolumn{1}{l|}{1}   & \multicolumn{1}{l|}{233}    & \multicolumn{1}{l|}{14}  & 15  \\ \cline{2-7} 
                        & times selected & \multicolumn{1}{l|}{78}   & \multicolumn{1}{l|}{78}   & \multicolumn{1}{l|}{70}    & \multicolumn{1}{l|}{3}  &   2 \\ \hline \hline
\multirow{2}{*}{4}      & variables      & \multicolumn{1}{l|}{2}   & \multicolumn{1}{l|}{1}   & \multicolumn{1}{l|}{233}    & \multicolumn{1}{l|}{94}  &  97  \\ \cline{2-7} 
                        & times selected & \multicolumn{1}{l|}{78}   & \multicolumn{1}{l|}{76}   & \multicolumn{1}{l|}{74}    & \multicolumn{1}{l|}{66}  &   2 \\ \hline \hline
\multirow{2}{*}{5}      & variables      & \multicolumn{1}{l|}{2}   & \multicolumn{1}{l|}{233}   & \multicolumn{1}{l|}{1}    & \multicolumn{1}{l|}{268}  &  36  \\ \cline{2-7} 
                        & times selected & \multicolumn{1}{l|}{78}   & \multicolumn{1}{l|}{37}   & \multicolumn{1}{l|}{32}    & \multicolumn{1}{l|}{29}  &   26 \\ \hline
\end{tabular}
\caption{
\label{tab:supernova_count}
The top five variables selected by the $M=78$ LOO CV for $\nnzero=1$--$5$. The result is similar to that of~\cite{obuchi2016sampling} using the SA-based method.
}
\end{center}
\end{table}

\section{Summary}\Lsec{Summary}
In this study, inspired by the theoretical result and the SA-based algorithm of~\cite{obuchi2018statistical,obuchi2016sparse,obuchi2016sampling}, we proposed an MC-based greedy algorithm called GMC for sparse linear regression. GMC is simpler than the SA-based algorithm but still it can achieve the perfect reconstruction in undersampling situations of a reasonable level, as shown by the numerical experiments on synthetic datasets. These experiments also suggest that GMC can outperform the $\ell_1$ relaxation which is the most commonly used method for sparse estimation. An additional experiment on a real-world dataset of supernovae also supported the practicality of GMC. These results imply that the energy landscape of the sparse linear regression problem is simple and exhibits a funnel-like structure~\cite{leopold1992protein} in the reconstructable region. We believe that this finding inspires further algorithms of sparse linear regression as well as new models for complex systems, such as glasses and proteins, based on sparse variable selection.

\section*{Acknowledgement}
This work was supported by JSPS KAKENHI Nos. 25120013, 17H00764, 18K11463 and 19H01812, JST CREST Grant Number JPMJCR1912, Japan. TO is also supported by a Grant for Basic Science Research Projects from the Sumitomo Foundation. 


\bibliographystyle{jpsj}
\bibliography{obuchi}

\begin{thebibliography}{10}

\bibitem{candes2005decoding}
E.~J. Cand{\`e}s and T.~Tao: IEEE transactions on information theory {\bfseries
  51} (2005) 4203.

\bibitem{candes2006robust}
E.~J. Cand{\`e}s, J.~Romberg, and T.~Tao: IEEE Transactions on information
  theory {\bfseries 52} (2006) 489.

\bibitem{candes2006near}
E.~J. Cand{\`e}s and T.~Tao: IEEE transactions on information theory {\bfseries
  52} (2006) 5406.

\bibitem{donoho2006compressed}
D.~L. Donoho: IEEE Transactions on information theory {\bfseries 52} (2006)
  1289.

\bibitem{natarajan1995sparse}
B.~K. Natarajan: SIAM journal on computing {\bfseries 24} (1995) 227.

\bibitem{pati1993orthogonal}
Y.~C. Pati, R.~Rezaiifar, and P.~S. Krishnaprasad: Signals, Systems and
  Computers, 1993. 1993 Conference Record of The Twenty-Seventh Asilomar
  Conference on, 1993, pp. 40--44.

\bibitem{davis1994adaptive}
G.~M. Davis, S.~G. Mallat, and Z.~Zhang: Optical engineering {\bfseries 33}
  (1994) 2183.

\bibitem{chartrand2008iterative}
R.~Chartrand and W.~Yin: Iterative Reweighted Algorithms for Compressive
  Sensing (2008).

\bibitem{tibshirani1996regression}
R.~Tibshirani: Journal of the Royal Statistical Society. Series B
  (Methodological)  (1996) 267.

\bibitem{meinshausen2004consistent}
N.~Meinshausen and P.~B{\"u}hlmann: 2004.

\bibitem{banerjee2006convex}
O.~Banerjee, L.~E. Ghaoui, A.~d'Aspremont, and G.~Natsoulis: Proceedings of the
  23rd international conference on Machine learning, 2006, pp. 89--96.

\bibitem{friedman2008sparse}
J.~Friedman, T.~Hastie, and R.~Tibshirani: Biostatistics {\bfseries 9} (2008)
  432.

\bibitem{donoho2009observed}
D.~Donoho and J.~Tanner: Philosophical Transactions of the Royal Society of
  London A: Mathematical, Physical and Engineering Sciences {\bfseries 367}
  (2009) 4273.

\bibitem{kabashima2009typical}
Y.~Kabashima, T.~Wadayama, and T.~Tanaka: Journal of Statistical Mechanics:
  Theory and Experiment {\bfseries 2009} (2009) L09003.

\bibitem{krzakala2012probabilistic}
F.~Krzakala, M.~M{\'e}zard, F.~Sausset, Y.~Sun, and L.~Zdeborov{\'a}: Journal
  of Statistical Mechanics: Theory and Experiment {\bfseries 2012} (2012)
  P08009.

\bibitem{kirkpatrick1983optimization}
S.~Kirkpatrick, C.~D. Gelatt, and M.~P. Vecchi: science {\bfseries 220} (1983)
  671.

\bibitem{obuchi2018statistical}
T.~Obuchi, Y.~Nakanishi-Ohno, M.~Okada, and Y.~Kabashima: Journal of
  Statistical Mechanics: Theory and Experiment {\bfseries 2018} (2018) 103401.

\bibitem{nakanishi2016sparse}
Y.~Nakanishi-Ohno, T.~Obuchi, M.~Okada, and Y.~Kabashima: Journal of
  Statistical Mechanics: Theory and Experiment {\bfseries 2016} (2016) 063302.

\bibitem{obuchi2016sparse}
T.~Obuchi and Y.~Kabashima: Journal of Physics: Conference Series, Vol. 699,
  2016, p. 012017.

\bibitem{obuchi2016sampling}
T.~Obuchi and Y.~Kabashima: 2016 24th European Signal Processing Conference
  (EUSIPCO), 2016, pp. 1247--1251.

\bibitem{silverman2012berkeley}
A.~V. Filippenko, M.~Ganeshalingam, W.~Li, and J.~M. Silverman: Monthly Notices
  of the Royal Astronomical Society {\bfseries 425} (2012) 1889.

\bibitem{Berkeley}
{The SNDB: http://heracles.astro.berkeley.edu/sndb/info}.

\bibitem{uemura2015variable}
M.~Uemura, K.~S. Kawabata, S.~Ikeda, and K.~Maeda: Publications of the
  Astronomical Society of Japan {\bfseries 67} (2015).

\bibitem{obuchi2016cross}
T.~Obuchi and Y.~Kabashima: Journal of Statistical Mechanics: Theory and
  Experiment {\bfseries 2016} (2016) 53304.

\bibitem{kabashima2016approximate}
Y.~Kabashima, T.~Obuchi, and M.~Uemura: 2016 54th Annual Allerton Conference on
  Communication, Control, and Computing (Allerton), Sep. 2016, pp. 596--600.

\bibitem{igarashi2018exhaustive}
Y.~Igarashi, H.~Takenaka, Y.~Nakanishi-Ohno, M.~Uemura, S.~Ikeda, and M.~Okada:
  Journal of the Physical Society of Japan {\bfseries 87} (2018) 044802.

\bibitem{igarashi2018es-dos:}
Y.~Igarashi, H.~Ichikawa, Y.~Nakanishi-Ohno, H.~Takenaka, D.~Kawabata,
  S.~Eifuku, R.~Tamura, K.~Nagata, and M.~Okada: Journal of Physics: Conference
  Series, Vol. 1036, 2018, p. 012001.

\bibitem{leopold1992protein}
P.~E. Leopold, M.~Montal, and J.~N. Onuchic: Proceedings of the National
  Academy of Sciences {\bfseries 89} (1992) 8721.

\end{thebibliography}

\end{document}